\documentclass[12pt,twoside]{article}   %For printing on two sides of page
\usepackage[super,sort,comma]{natbib}
\usepackage{fancyhdr}		%Gives headers and footers defined below
\usepackage[section]{placeins}   %
\usepackage{graphicx}

\makeatletter \renewcommand\@biblabel[1]{$^{#1}$} \makeatother
\newcommand{\cen}[1]{\begin{center} #1 \end{center}}

\usepackage[mathlines]{lineno}
\usepackage{hyperref}
\hypersetup{ colorlinks,
    citecolor=blue,
    filecolor=blue,
    linkcolor=blue,
    urlcolor=blue
}

\usepackage{xcolor}
\definecolor{gray}{rgb}{0.6,0.6,0.6}
\definecolor{red}{rgb}{0.85,0,0}
\definecolor{green}{rgb}{0,0.85,0}
\definecolor{blue}{rgb}{0,0,0.85}
\definecolor{beige}{rgb}{0.92,0.87,0.78}
\usepackage[all]{hypcap}    %causes link to figures to go to figure, not caption
\usepackage{multirow}
\usepackage{caption}
\usepackage{subcaption}
\usepackage{chngcntr}
\usepackage{xr}

\begin{document}

\cen{\sf {\Large {\bfseries Comparing the Performance of Foundation Model Derived Embeddings with Traditional Approaches for Distant Metastasis Prediction in Head and Neck Cancer} \\  
\vspace*{10mm}
Erich Schmitz, Meixu Chen, Bowen Jing, Jing Wang} \\
Advanced Imaging and Informatics for Radiation Therapy (AIRT) and Medical Artificial Intelligence and Automation (MAIA) Laboratory, Department of Radiation Oncology, University of Texas Southwestern Medical Center, Dallas, Texas
\vspace{5mm}\\
Version typeset \today\\
}

\pagenumbering{roman}
\setcounter{page}{1}
\pagestyle{plain}
\textbf{Corresponding Author:} Jing Wang \\
\indent \textbf{E-mail:} jing.wang@utsouthwestern.edu
% note, probably best not to use a student's e-mail as it won't be valid for
% very long.

\begin{abstract}
\noindent {\bf Background:} Early prediction of distant metastasis (DM) risk in head and neck cancer (HNC) can enable timely interventions that may improve treatment outcomes. While machine learning approaches using medical imaging have been widely explored for this task, many current methods rely on prior knowledge of the region of interest such as tumor segmentations, which require expert knowledge, is time-consuming and introduces user-dependent variability. Medical image-based foundation models have recently been developed for specific imaging modalities to streamline down-stream prediction tasks by extracting modality-relevant features.\\ 
{\bf Purpose:} In this study, we evaluate the effectiveness of using a foundation model as the feature extractor from minimally processed image volumes to predict DM risk in HNC patients and compare its performance with traditional approaches that require prior knowledge on the regions of interest.\\
{\bf Methods:} Preoperative CT images of 2327 patients from the RADCURE dataset were used. Three features-sets were created including radiomics, deep-learning based features, and CT Foundation derived features. The feature-sets were used individually in a multi-layer perceptron (MLP) to predict DM risk. The final models for each feature-set were chosen using 5-fold cross validation and their performance was evaluated using a hold-out testing set.\\
{\bf Results:} The model using CT Foundation embeddings outperformed the radiomics and deep learning-based models, achieving a Receiver Operating Characteristic Area Under the Curve (AUC) of 0.791, compared to AUC values of 0.772 and 0.753 for the radiomics and deep learning-based models, respectively. The CT Foundation based model had similar performance to a model that combined the use of radiomics and deep learning-based features that achieved an AUC of 0.794. \\
{\bf Conclusions:} Features based on foundation models offer a promising alternative to traditional radiomics while reducing the need for domain expertise and extensively annotated datasets. Their minimal preprocessing requirements also make them a more accessible and scalable option.   \\

\end{abstract}

\newpage     %may or may not be needed

%The table of contents is for drafting and refereeing purposes only. Note
%that all links to references, tables and figures can be clicked on and
%returned to calling point using cmd[ on a Mac using Preview or some
%equivalent on PCs (see View - go to on whatever reader).
\tableofcontents

\newpage

\setlength{\baselineskip}{0.7cm}      %double spacing		

\pagenumbering{arabic}
\setcounter{page}{1}
\pagestyle{fancy}
\section{Introduction}

Head and neck cancer (HNC) is the sixth most common cancer in the world with a 5-year survival of 66\% \citep{RN1}. Treatments commonly include radiation therapy with the possibility of concurrent or adjuvant chemotherapy. The form of treatment is determined from diagnostic imaging and health assessments that will inform the physician of the severity of the cancer and the risk of recurrence. For HNC there are three main factors in determining risk: the state of the primary tumor, spread to the lymphatic system, and the occurrence of distant metastasis (DM) \citep{RN2}. For patients undergoing definitive treatment, early prediction of DM after initial diagnosis plays a critical role in clinical management. Identifying patients at high risk for DM allows for timely intervention with intensified systemic therapy, which may reduce the likelihood of metastasis and ultimately improve outcomes such as overall survival.

Machine learning and artificial intelligence-based imaging analyses have been widely explored for patient prognosis \citep{RN3, RN4}. A common approach to these tasks is to extract relevant features from medical images and use these features to perform a prediction. Common methods of feature extraction include the creation of handcrafted features, such as with radiomics, or directly learning a set of features using a deep neural network \citep{RN3, RN5, RN6}. Previous studies have made use of these methods in outcome prediction tasks for HNC and have demonstrated their value \citep{RN4, RN7, RN8, RN9}. However, both handcrafted feature-based and deep learning-based approaches require prior knowledge of the regions of interest (ROIs) to be analyzed. 

Specifically, radiomics requires accurate contouring of the ROI, which can be determined manually by experts or automatically through an automated segmentation algorithm. However, both approaches have their limitations: manual delineation can be time-consuming and subjective, while the use of limited and potentially biased reference contours for training can compromise the reliability and accuracy of automated segmentations. Deep learning-based methods also generally require determination of an ROI, though not as stringent as for radiomics, only requiring a bounding box. This bounding box necessitates the knowledge of the location of a tumor or ROI. Additionally, for both feature extraction methods, the localized nature of the ROIs will ignore prognostic information from surrounding tissue that could be useful in a prediction task. In a clinical setting, ROI-dependence can introduce bias based on subjective ROI-delineation which has the potential to affect patient outcomes. Additionally, the need for the delineation takes time with limited expertise available, potentially increasing the time to treatment and affecting the lives of physicians through increased after-hours workloads \cite{RN159}.

Other studies have been performed that either make use of whole-volume CT images or try to remove the need for a ROI or segmentation. These studies include tasks such as abnormality prediction and detection, automated segmentation, image quality enhancement, and survival/outcome prediction \cite{RN96, RN152, RN153}. Some remove the need for a ROI entirely, while others define a general ROI while not including segmentations. For those that deal with survival prediction, one emphasizes the removal of any manual delineation by using a pre-trained deep learning-based feature extractor on 2D projections of PET images \cite{RN96}. A second study used a transfer-learning based approach on CT volumes, removing the need for precise segmentation, but still requiring an ROI that encompasses the tumor volume, tumor edges and adjacent tissue \cite{RN155}. Their pre-trained feature extractor, which didn't use segmentations, outperformed the radiomics-based model. 

Recently, several medical image-based foundation models have been developed, where models trained on large-scale datasets from specific imaging modalities are able to perform a wide range of machine learning tasks \citep{RN10, RN11, RN12, RN13, RN14}. These foundation models are domain specific, enabling medical imaging tasks to focus on modality-relevant features, and can serve as pre-trained feature extractors.  One of these foundational models is called CT Foundation, a model that was created with the purpose of reducing image processing complexity and removing the need for expert curation of a dataset. CT Foundation is a tool that directly processes CT volumes, providing a 1-dimensional embedding vector to summarize the features of the volume \citep{RN11, RN15}. It was developed with the intention of using the embeddings as input to downstream classifiers and so requires minimal data processing in its use. As it processes the full CT volume, this model does not require the delineation of an ROI, giving it the potential to overcome some of the limitations in using radiomics and individually produced feature extractors.  Additionally, since the images are converted to 1-dimensional embeddings, much like with the use of radiomics, the computational burden of training and fine-tuning a 3D network is removed allowing a quicker turnaround for model training and testing.

This study aims to evaluate the effectiveness of CT foundation models for predicting DM in HNC patients. To this end, we compare the predictive performance of foundation model-derived embeddings with more traditional approaches, including radiomics and deep learning based features. Through these comparisons, we seek to investigate whether foundation models can achieve comparable or superior performance while reducing the computational burden and expert input typically required for radiomics-based analysis and 3D image processing.

\section{Methods and Materials}

\subsection{Dataset}

This study utilized a public dataset RADCURE downloaded from TCIA  \citep{RN16}. The RADCURE dataset is a cohort of 3346 HNC patients treated with radiation or chemoradiation therapy at the Princess Margaret Cancer Center in Toronto, Canada \citep{RN17}. The dataset contains CT images created using standard imaging protocols, with 61.5\% of the patients being administered contrast agents. Contours are provided for the primary and lymph node gross tumor volumes (GTVs). The contours were manually delineated according to the cancer center’s guidelines, and peer-reviewed during quality assurance rounds. In addition to the images, patient outcomes and clinical features are provided. These features include demographics, tumor staging, treatment information and whether the patients were supplied with contrast.

Patient selection included requiring the existence of a primary GTV (GTVp) and removing a sub-set of censored patients for which their last follow-up time was within 2-years without distant failure. This resulted in 2327 patients, with 375 having distant failures within 2 years. The patients were split into training sets, with 1819 allocated to the training/validation set and 508 to the test set. The test set was chosen to match that used in the RADCURE challenge \citep{RN8}, to allow for easier and direct comparisons. Table \ref{tab:patients} summarizes demographics and outcomes of the selected patient set. 

\begin{table}[h!]
	\vspace{1em}
	\centering 
	%\addtolength{\leftskip}{-2cm}
	%\addtolength{\rightskip}{-2cm}
	\begin{tabular}{c c c c c}
		\hline
		\multicolumn{2}{c}{Feature} & Total & Training/validation & Test  \\
		\hline
		\multicolumn{2}{c}{Patient Total} & 2327 & 1819 & 508 \\
		\multicolumn{2}{c}{2-year Distant Failure} & 375 (12\%) & 291 (16\%) & 84 (17\%) \\
		Sex & & & & \\
		 & Male & 1873 (81\%) & 1453 (80\%) & 420 (83\%) \\
		 & Female & 454 (19\%) & 366 (20\%) & 88 (17\%) \\
		\multicolumn{2}{c}{Age} & 15-90 (IQR: 53.9-69.4) & 15-90 (IQR: 53.8-69.5) & 22-90 (IQR: 54.7-68.9) \\
		\hline	
	\end{tabular}
	\caption{Patient demographics, including the occurrence of 2-year distant failure. The patients numbers are divided into the training and testing sets. The percentages are with regards to the row ‘Patient Total’ for the relevant column. For age, the interquartile range (IQR) is given in addition to the full range.}
	\vspace{1em}
	\label{tab:patients}
\end{table}

\subsection{Data preparation and feature extraction}

The original CT images were preprocessed according to the requirement of different feature extractors. These corresponded to CT foundation embeddings, radiomics and a Vision Transformer (ViT) network that was trained from scratch. Before the individual processing, the images were converted form DICOM to the NifTI file format.  The CT foundation model processes the full CT volume, where, for this study, the individual images were reduced axially, in order to save space when uploading the images to the cloud server. This reduction was done following the location of the tumor volume to prevent clipping off parts of the tumor, while keeping the 512x512 slices unprocessed. For the radiomics the full CT volumes with their GTVp contours were provided. For the ViT, some standard image preprocessing was applied, including resampling to a consistent 1x1x1 $mm^3$ spacing, cropping to a size of 80x80x80 voxels around the center of the GTVp, truncating the HU values to a range [-500, 500] and subsequently normalizing the image to a range of [-1, 1] using Min-Max normalization.

The CT foundation embeddings were produced by using the CT volumes as input into the CT Foundation API \citep{RN11}. The CT Foundation uses a contrastive captioner (CoCa) model that takes CT volumes paired with radiology reports and encodes them into a shared embedding space. The foundation model works by minimizing two types of loss: a captioning loss for the reports and a contrastive loss to improve the semantic understanding of the images \cite{RN15}. The model was trained on over 500,000 multi-institutional CT images with a wide range of body parts including the head and extremities \cite{RN15}. The process output a set of 1408 features to be used as input for downstream tasks. The embeddings were created following the notebooks in the official repository provided by the creators of the CT Foundation model \cite{RN158}. 

The radiomics features were extracted by pyradiomics \citep{RN18}. A total of 1316 features were extracted including the first order, shape, gray level cooccurrence matrix, gray level run length matrix, gray level size zone matrix, neighboring gray tone difference matrix and gray level dependence matrix classes. These features were obtained from 6 derivative filters in addition to the original: wavelet, square, square root, logarithmic, exponential, and gradient filters.

The third feature extractor was a vision transformer (ViT) that was trained from scratch with the multi-layer perceptron that was used as the classification arm. The image patches previously mentioned were used as input to the ViT to produce a set of 512 features. The ViT was initialized using Monai with default parameters except for the patch size set at 16, the image size set at 80x80x80, and the number of layers set to 10.  

Additionally, 14 clinical features were used as additional input to the model during the training, which were concatenated to the feature vectors before the classification step. These clinical features included age, sex, ECOG, smoking status, smoking packs per year, Cancer site, T and N stage, AJCC Stage 7th edition, HPV status, planned treatment modality, dose, number of fractions, and the presence of contrast. The categorical features were assigned integers values according to their status. For features with binary categories values of -1 and 1 were given, with 0 reserved for undefined or missing values. T, N, AJCC Stage, and ECOG were given integer values in the order of their stage number. A value of -1 was given to undefined stages and other values ranged from 0 to 5 for T-stage, 6 for N-stage, 9 for AJCC Stage, and 4 for ECOG. For T-stage the max value was given to those labeled with T4b, for N-stage N3b, for AJCC IVC, and for ECOG, ECOG 4. Apart from the HPV status, the number of patients with undefined clinical variables is less than 1\% of the total, so imputation was not considered

\subsection{Model Architecture}

Two types of classification models were used for feature set comparisons. One model type was a deep neural network (DNN) made from a 4-layer multi-perceptron (MLP), while the other was a traditional ML model using a support vector machine (SVM). The MLP models made use of the three sets of features as input during training, while the SVM models only made use of the foundation and radiomics based features. The MLP classifier had a total of four dense layers, with two of them hidden. The first three layers used a ReLU activation function and the final layer used a Sigmoid activation function. The number of channels in the hidden layers were set to 2, after testing different combinations of up to 512 channels. The clinical features were concatenated to the output after the third layer, but before the final layer. The networks and subsequent training were implemented using the pytorch and pytorch-lightning frameworks \citep{RN22, RN23}. The SVM classification models used a feature selection algorithm followed by the SVM. For feature selection a minimum Redundancy Maximum Relevance algorithm was used \citep{RN162}. 

\begin{figure}
	\vspace{1em}
	\centering
	\includegraphics[width=\textwidth]{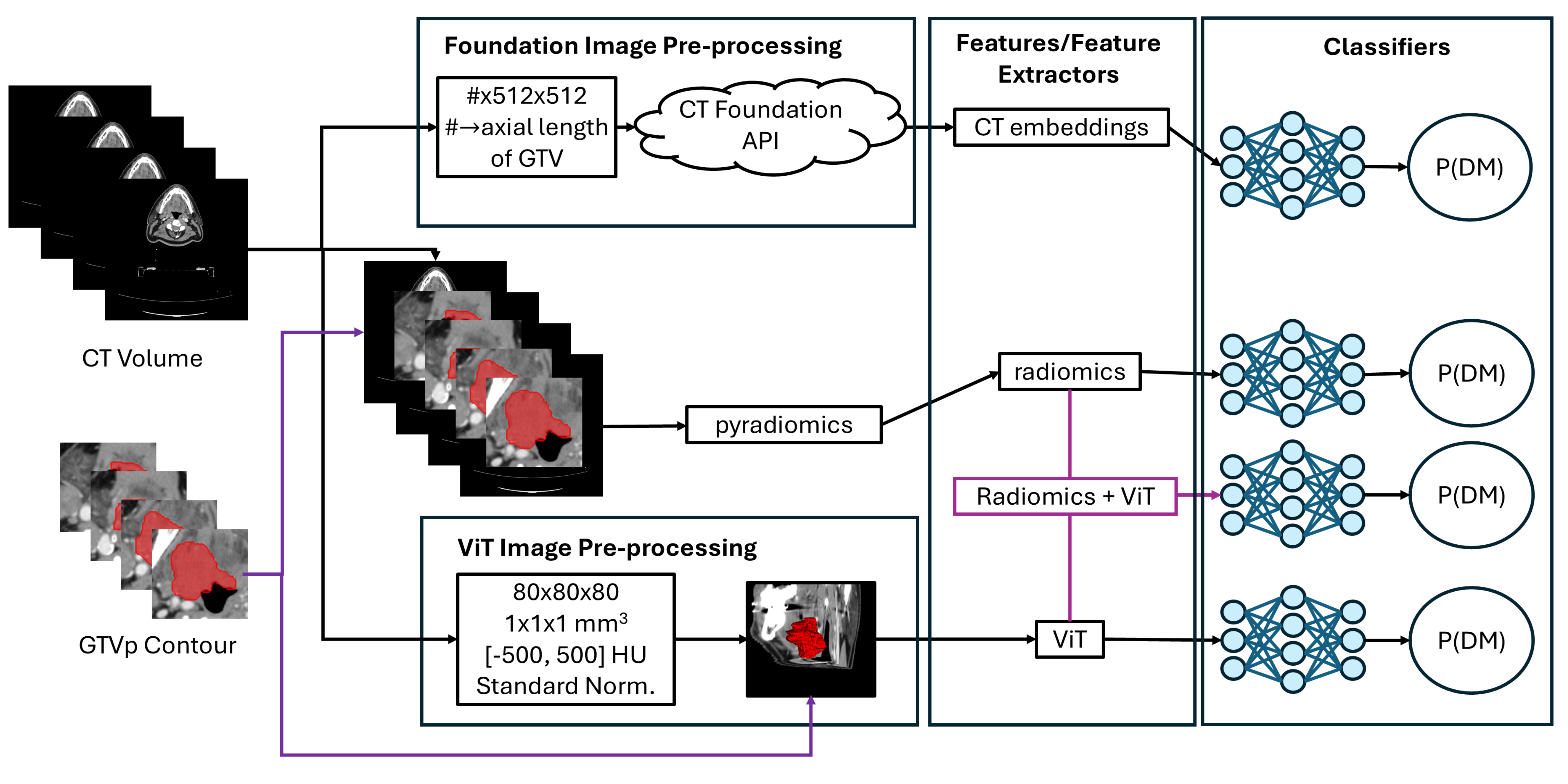}
	\caption{A diagram showing input for the models. For the CT foundation embeddings, the only input is an axial section of the CT volume. For the radiomics and ViT features, the input includes both the CT volume and GTVp contours, with image preprocessing applied for the ViT arm to pull out the relevant image ROI and to apply normalizations. The box labeled Classifiers can refer to an MLP for all inputs or an SVM for the CT foundation embeddings or radiomics.}
	\label{fig:1}
	\vspace{1em}
\end{figure}

A total of seven models were produced, with their differences corresponding to the feature inputs and whether an MLP or SVM were used as the classifier. A diagram of the workflow is illustrated in \ref{fig:1}. There were four sets of extracted features used as inputs (\ref{fig:1}), corresponding to CT embeddings, radiomics, ViT, and a combination of radiomics and ViT features.  The inclusion of this last combination of features was motivated by several previous studies showing the complimentary nature of handcrafted and learned features, where the combination of these two categories of features have led to improved model performance \citep{RN19, RN20, RN21}. The MLMP classifiers used all four of the feature sets, while the SVM classifiers only used the CT foundation and radiomics features, with an additional model that used the clinical features only for reference. All of the feature sets had standard normalization applied before being fed to the pipeline.

\subsection{Training and Evaluation}

The MLP models were trained using 5-fold cross validation and tested on the hold-out test set used in the RADCURE challenge.  The training lasted for 100 epochs and used a learning rate of 0.001 with an ADAM optimizer. The learning rate followed a plateau-based scheduler with a reduction factor of 0.1 and a patience of 10 epochs.  The hyperparameters were tuned using grid-based scans of the batch size, number of epochs, learning rate, and when the ViT is used, the number of channels output from the ViT.  From this parameter tuning, we determined the final set of parameters that the models were trained with.     

An ensemble of the models from the 5 CV folds was made by averaging the probabilities of the models together. This ensemble model was evaluated against the testing dataset, comparing models that used different feature input sets during the training. Performance of the individual folds are given in the supplementary material, Table S6. 

The SVM models were trained using a gridgrid search and 5-fold cross-validation. The cross-validation folds were the same as those used in the MLP model training.  The grid search was performed using the GridSearchCV function from the scikit-learn python package. The optimized parameters included the number of features selected from the mRMR algorithm,  and the C, gamma and kernel parameters from the SVM. The grid search found the best performing parameters for each set of features given in Table \ref{tab:parameters}. A table of the full grid search, along with a list of selected radiomics features and foundation embeddings can be found in Tables \ref{tab:supp_radiomics}, \ref{tab:supp_foundation}, and \ref{tab:supp_grid} of the supplemental materials. 

\begin{table}[h!]
	\vspace{1em}
	\centering 
	%\addtolength{\leftskip}{-2cm}
	%\addtolength{\rightskip}{-2cm}
	\begin{tabular}{c c c c}
		\hline
		Input & Radiomics & Foundation & Cli-only \\
		\hline
		mRMR & \multirow{2}{*}{5} & \multirow{2}{*}{30} & \multirow{2}{*}{-} \\
		Num. Selected Features & & & \\
		SVM C & 1& 1 & 1\\
		SVM gamma & 0.01 & 0.01 & 0.01 \\
		SVM kernel & rbf & rbf & rbf \\
		\hline	
	\end{tabular}
	\caption{SVM parameter selections for the traditional ML models that use Radiomics, CT foundation embeddings, and clinical features only.}
	\label{tab:parameters}
	\vspace{1em}
\end{table}

Four metrics were calculated as part of the evaluation: AUC, average precision (AP), sensitivity (SEN), and specificity (SPE). The AUC was used to gauge the overall performance, the AP to determine how effective the model was in correctly predicting the positive class, and the remaining two were used to gauge the balance of predictions to ensure that the AUC was not biased by the imbalance in the positive and negative classes. 

Predictions were recalibrated using Platt scaling, then thresholds (initially 0.5) were adjusted. For the MLP-based models, calibration was applied individually to out of fold (OOF) logits from each of the five models prior to ensembling. The calibration curve was then plotted using the ensemble of the newly calibrated predictions. For the SVM-based models, the predictions were calibrated with the CalibratedClassifierCV function from scikit-learn.  The threshold tuning was performed to determine a point that best balances the sensitivity and specificity, while prioritizing sensitivity, by minimizing the cost function given in Equation \ref{eq:cost}. The new thresholds for each model are given in Table \ref{tab:supp_threshold} of the Supplemental materials. Calibration curves before and after calibration are also provided in Figures \ref{fig:supp_calibration_dnn}, \ref{fig:supp_calibration_trad}.

\begin{equation}
	cost = \left(1-Sen\right) + \left|Sen-Spe\right|
	\label{eq:cost}
\end{equation}

Statistical comparisons were made using the R language. P-values were calculated using a one-sided DeLong test for two correlated ROC curves \citep{RN24}. The p-values were corrected for multiple comparisons using the False Discovery Rate (FDR) method.

\section{Results}

\begin{figure}
	\centering
	\includegraphics[width=\textwidth]{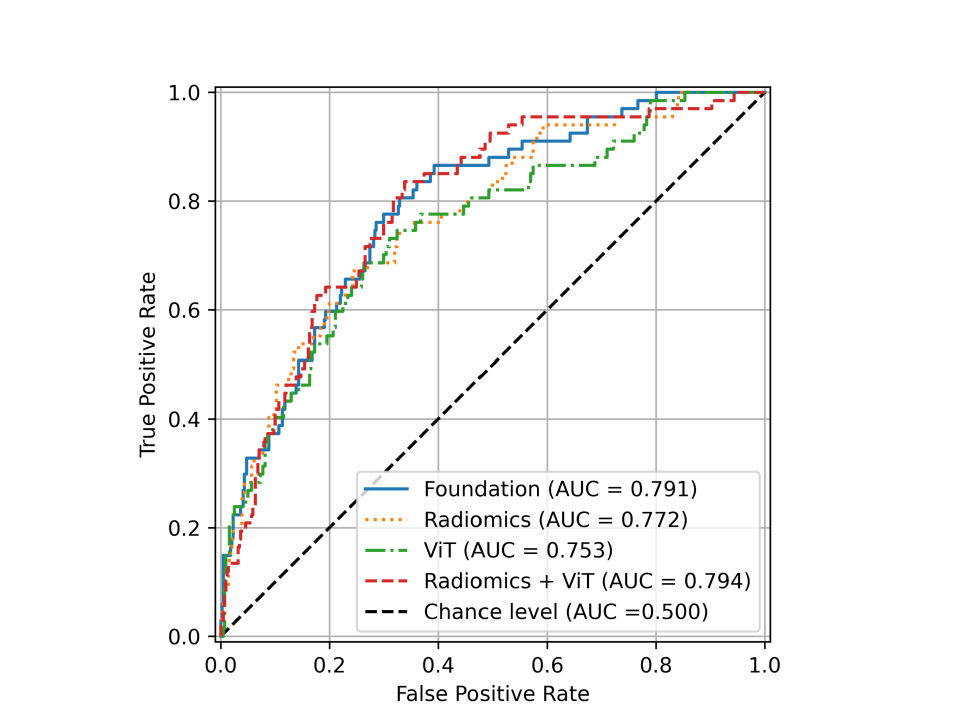}
	\caption{ROC curves of the SVM-based models, corresponding to using radiomics, foundation embeddings, and clinical features only.}
	\label{fig:2}
\end{figure}

\begin{figure}
	\centering
	\includegraphics[width=\textwidth]{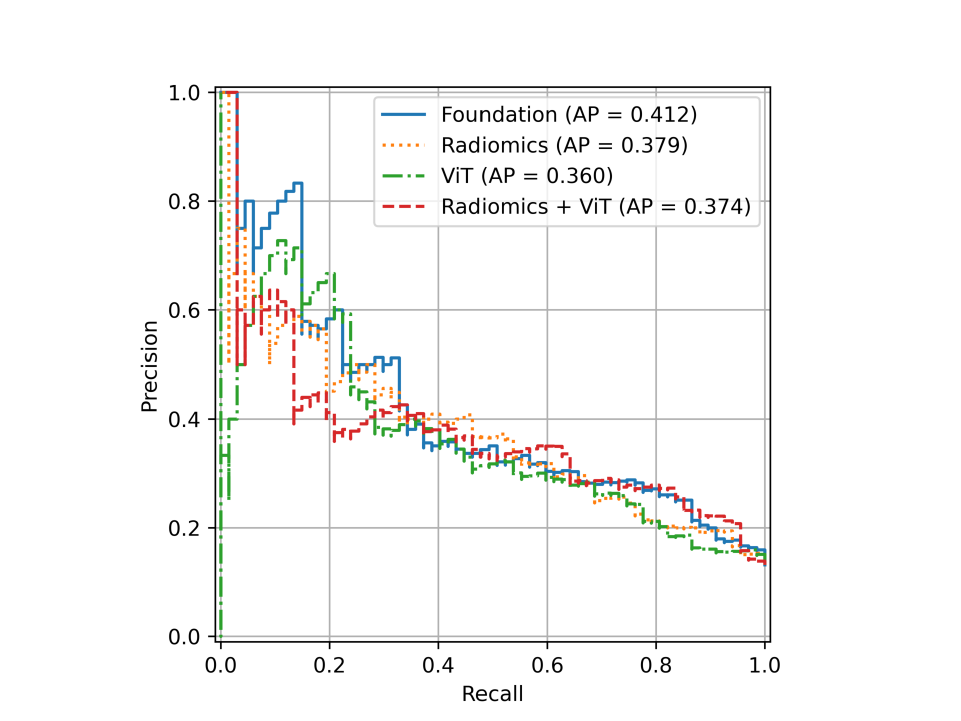}
	\caption{PR curves curves of the SVM-based models, corresponding to using radiomics, foundation embeddings, and clinical features only.}
	\label{fig:3}
\end{figure}

\begin{figure}
	\centering
	\includegraphics[width=\textwidth]{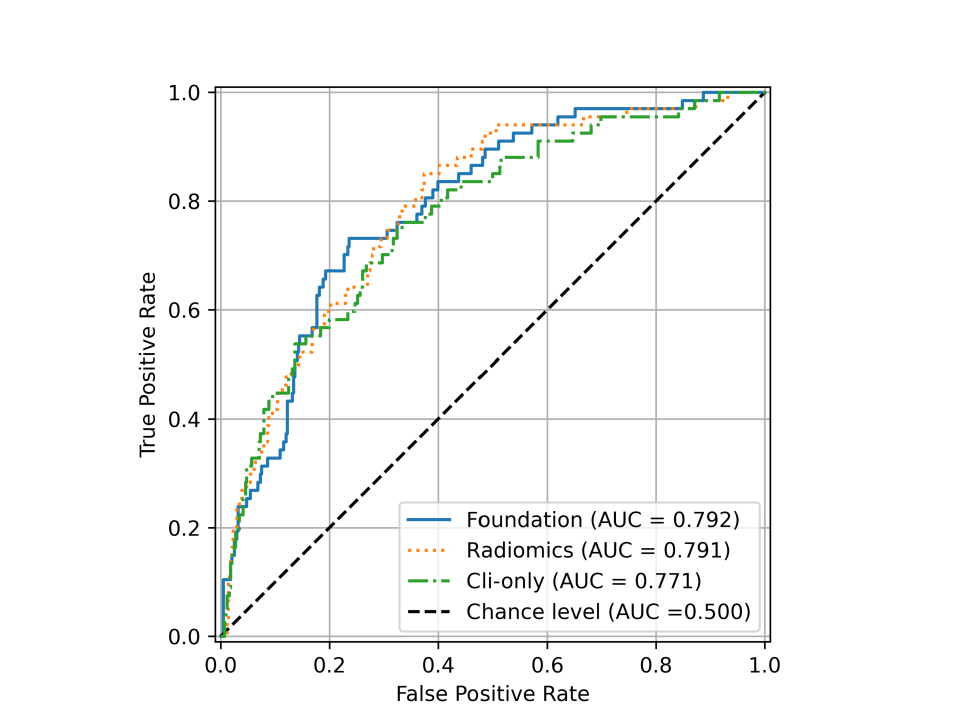}
	\caption{ROC curves of the MLP-based models using 4 different sets of inputs corresponding to CT Foundation embeddings, radiomics, ViT extracted features, and radiomics combined with ViT extracted features.}
	\label{fig:4}
\end{figure}

\begin{figure}
	\centering
	\includegraphics[width=\textwidth]{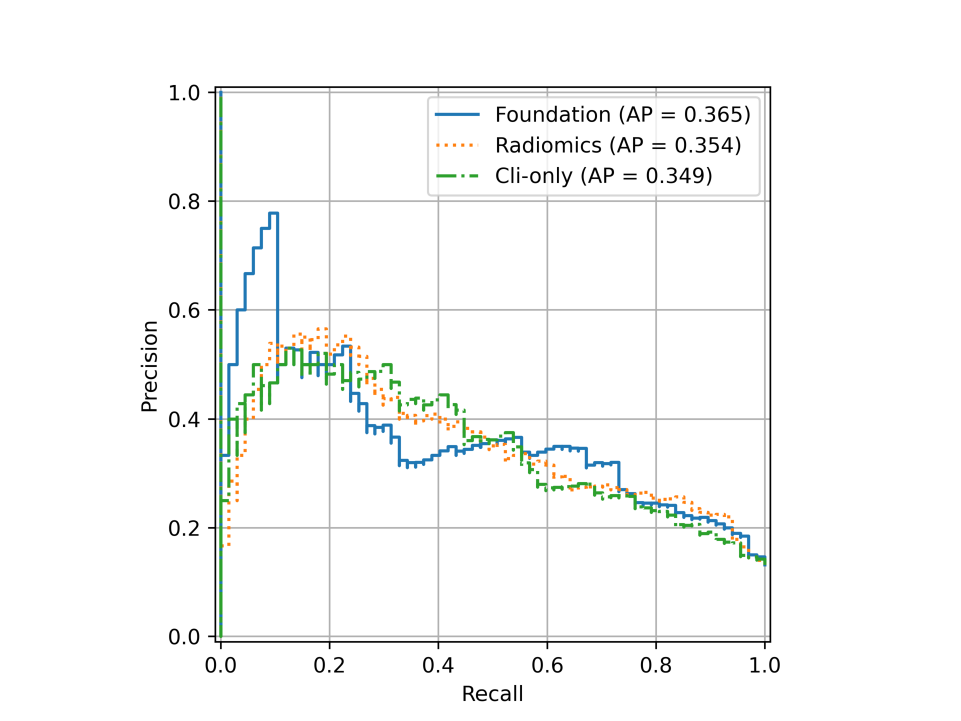}
	\caption{PR curves of the MLP-based models using 4 different sets of inputs corresponding to CT Foundation embeddings, radiomics, ViT extracted features, and radiomics combined with ViT extracted features. }
	\label{fig:5}
\end{figure}

The model comparisons focus on evaluating the use of CT foundation embeddings—which do not rely on the full GTVp contours—against more traditional approaches that incorporate GTVp contours, such as radiomics and cropped image regions surrounding the contour. Emphasis is placed on the performance of the MLP-based models, given in Table \ref{tab:results}, to compare the performance of the four feature sets within the use of one type of classifier. Additional comparisons of the performance of the SVM-based models are provided in Table \ref{tab:trad_results}. ROC and PR curves for the MLP models are given in Figures \ref{fig:2} and \ref{fig:3}, respectively. The same plots are given for the SVM models in Figures \ref{fig:4} and \ref{fig:5}.  

\begin{table}[h!]
	\vspace{1em}
	\centering 
	%\addtolength{\leftskip}{-2cm}
	%\addtolength{\rightskip}{-2cm}
	\begin{tabular}{c c c c c}
		\hline
		Extracted feature source & AUC $[95\%CI]$ & AP & SEN & SPE \\
		\hline
		Foundation & \textbf{0.791} $\mathbf{[0.736,0.846]}$ & \textbf{0.412} & \textbf{0.687} & \textbf{0.730} \\
		Radiomics & 0.772 $[0.712, 0.832]$ & 0.379 & 0.687 & 0.712 \\
		ViT & 0.753 $[0.688, 0.817]$ & 0.360 & 0.687 & 0.707 \\
		Radiomics + ViT & \textbf{0.794} $\mathbf{[0.739, 0.849]}$ & \textbf{0.374} & \textbf{0.761} & \textbf{0.687} \\
		\hline	
	\end{tabular}
	\caption{Results of the MLP classifier with different inputs: radiomics, CT foundation embedding, ViT extracted features, and a combination of radiomics and ViT features. The AUC, AP, SEN, and SPE are shown for the 5-fold ensemble. The 95\% confidence intervals are given for the AUC of the ensemble models.}
	\vspace{1em}
	\label{tab:results}
\end{table}

\begin{table}[h!]
	\vspace{1em}
	\centering 
	%\addtolength{\leftskip}{-2cm}
	%\addtolength{\rightskip}{-2cm}
	\begin{tabular}{c c c c c}
		\hline
		ML Model Input & AUC $[95\%CI]$ & AP & SEN & SPE \\
		\hline
		\textbf{Foundation} & \textbf{0.791} $\mathbf{[0.737,0.846]}$ & \textbf{0.365} & \textbf{0.866} & \textbf{0.522} \\
		\textbf{Radiomics} & \textbf{0.792} $\mathbf{[0.737, 0.847]}$ & \textbf{0.354} & \textbf{0.896} & \textbf{0.537} \\
		Clinical-only & 0.771 $[0.711, 0.832]$ & 0.349 & 0.836 & 0.535 \\
		\hline	
	\end{tabular}
	\caption{Results of the SVM classifier with radiomics, CT foundation embeddings, or clinical features only as input. The AUC, AP, SEN, and SPE are shown for the final model.  The 95\% confidence intervals are given for the AUC.}
	\vspace{1em}
	\label{tab:trad_results}
\end{table}

The performance of the MLP models over individual folds and for some select patient populations (stratified by sex, with and without contrast, as well as with and without clinical features) are presented in Tables \ref{tab:supp_folds}, \ref{tab:supp_strat} of the Supplement for the radiomics and CT foundation models. Subgroup analyses compared contrast-enhanced versus non-contrast images and male versus female patients. Models trained without clinical features were also evaluated, showing that clinical variables improved performance and help mitigate population-related bias. 

The MLP model for the Foundation Embedding input showed the best performance among the three individual feature inputs with an AUC of 0.791 compared to the models using radiomics and ViT features with AUCs of 0.772 and 0.753, respectively.  The combination of radiomics and ViT features had the best overall performance, with an AUC of 0.794, but compared to the results of the Foundation embedding model there is not a significant difference, with a corrected p-value of 0.558. Of particular note with these two models is the increase in AP when moving from the radiomics+VIT model to the Foundation based model, indicating that the Foundation model may have a better potential in reducing the number of false positives. The p-values of the MLP models, as given in Table \ref{tab:pvalues}, show that there is a statistical significance with $p<0.05$ between the Foundation embedding model and the ViT feature model, if one does not consider p-value corrections for multiple comparisons, otherwise no significant differences were observed. 

\begin{table}[h!]
	\vspace{1em}
	\centering 
	%\addtolength{\leftskip}{-2cm}
	%\addtolength{\rightskip}{-2cm}
	\begin{tabular}{c c c c}
		\hline
		Model 1 & Model 2 & Initial p-value & corrected p-value \\
		\hline
		\multirow{3}{*}{Foundation} & Radiomics & 0.190 & 0.285 \\
		& ViT & 0.030 & 0.090\\
		& Radiomics + ViT & 0.442 & 0.558 \\
		\hline	
	\end{tabular}
	\caption{P-values comparing MLP based models, calculated using a one-sided DeLong’s test for correlated ROC curves. The initial p-values and those corrected for multiple comparisons are given.}
	\vspace{1em}
	\label{tab:pvalues}
\end{table}

For the models trained using traditional machine learning, the SVM models showed a comparable performance to their MLP counterparts, except for the radiomics model, which showed an improved performance in the AUC compared to the MLP version. The major performance difference between the SVM and MLP based models is the balance of sensitivity and specificity, and by extension their ability to handle the class imbalance present in the dataset. The MLP models gave SEN/SPE ratios ~0.9 showing a better balance compared to the SVM models which gave ratios of ~1.8, even after applying calibration and threshold tuning. It should be noted that the calibration and threshold tuning did not have a significant effect on the SEN/SPE ratio for the MLP models, while the opposite is true for the SVM models, which had more severe imbalances favoring the specificity before tunning. This difference can also be seen in the AP, with the MLP models showing higher values for models of similar performance in terms of AUC.   

\section{Discussion and Conclusion}

In this study, we explored the utility of CT foundation embeddings as an alternative to traditional radiomics and ViT-based feature extraction, with the goal of reducing the data burden and expert-driven processing typically required for model development. We generated three distinct feature sets—CT foundation embeddings from CT images where the axial slices had no cropping applied, radiomics features from masked tumor images, and leaned features from bounding boxes—and trained models on each using a simple linear network to isolate the influence of input features on prediction performance. For the specific task of predicting DM for HNC after definitive chemoradition or radiation therapy, our results demonstrate that the foundation embeddings outperformed or where comparble to radiomics alone, ViT-based features and radiomics+ViT inputs. Notably, the CT foundation embeddings do not require prior contouring of the GTVp, offering a significant advantage in terms of scalability and automation. This highlights the potential of foundation models to streamline predictive modeling pipelines by eliminating the need for time-consuming, expertise-dependent steps such as manual segmentation. Since the foundation model is pretrained on a large, annotated dataset, it does not require a large dataset on the side of the researcher, allowing for more meaningful results with smaller datasets. In addition, compared to training on images, using the CT foundation embeddings greatly reduces the number of trainable parameters by nearly 5 orders of magnitude, from 97 million parameters to 2600.  This makes the downstream training simple enough to efficiently run on a CPU, similar to radiomics only models, but processing the whole image, with a reduced need for physician defined contours or ROIs.

Some detractions from the use of the CT Foundation embeddings are related to the current ease of access to the API. As it is a cloud-based API, it requires sufficient permissions including storage, file-access and API access. Due to its novelty, it also cannot yet be considered user-friendly, with current documentation amounting to examples of the full process to produce the embeddings. While these examples are helpful, they are not one-size fits all and may require some experimentation to produce a proper workflow. Once the use of foundation models becomes more widespread it is foreseen that the ease of use will be improved. Another detraction is related to the storage and use of medical images in a cloud-based environment. To use the API it is necessary to have the relevant volumes uploaded into a container on the cloud, with those images also being processed in the cloud. For private datasets, this puts some onus on the researcher to ensure proper handling of protected health information, especially for those who are using a private account to access the API. On the other hand, since the processing is cloud-based, it does not require any local computing resources like gpu's to extract features, increasing its accessibility to places with smaller computational resources.

In this study, the deep learning–based ViT model trained from scratch was restricted to a tumor centered region of interest, whereas the CT foundation model processed larger portions of the CT volume. This design reflects practical and methodological considerations rather than an attempt to create asymmetric comparisons. Training a whole volume 3D neural network from scratch would substantially increase input dimensionality, memory consumption, and the number of trainable parameters, rendering such an approach computationally infeasible within our current framework. Moreover, prior work has suggested that tumor focused regions may capture more discriminative information for prediction than unsegmented whole volume inputs. In contrast, the CT foundation model leverages large scale pretraining to efficiently encode global contextual information without requiring extensive retraining or segmentation, representing a fundamentally different modeling paradigm. As such, the comparison highlights the practical trade offs between conventional tumor based deep learning pipelines and foundation model approaches that enable whole volume analysis with significantly reduced local computational burden.

In our cohort, male patients substantially outnumber female patients (~81\% vs. 19\%), which is representative of the sex distribution in this disease population and may influence model behavior. To assess potential bias, we performed sex stratified evaluations and observed similar AUCs for males and females across both the CT foundation– and radiomics based models, with modest differences in operating characteristics (e.g., lower sensitivity but higher specificity in the smaller female subgroup). These patterns likely reflect sampling variability and the reduced number of positive events among females, rather than systematic model bias. We have added the stratified results to Table S5 and note that future work will prioritize more balanced training cohorts and/or reweighting strategies to further mitigate sex related performance variability.

Potential limitations to this study include the use of a single institution dataset, along with only using CT Foundation as a foundation model, without considering other pre-trained cases, including other potential foundational models, such as the Vision Foundation Model for Computed Tomography (CT-FM) \citep{RN12}. Additionally, this study did not actively account for differences in the use of contrast in the CT images, only including the information as a binary variable in the clinical features concatenated with the set of extracted image features. Future work could include the comparison of multiple forms of pre-trained networks, including other foundation models and other forms of transfer learning.  Separate analyses of features extracted from contrast-enhanced and non-contrast-enhanced images could further clariy whether contrast status influences model performance. 

In conclusion, the reduced prior knowledge required to generate CT foundation embeddings highlights their potential as a strong alternative to traditional radiomics. This approach reduces the burden associated with medical image analysis, offering a more efficient and accessible solution for those without immediate access to domain expertise, extensive computing resources, or large annotated datasets. 

\section{Acknowledgment}

We acknowledge the funding support from the National Institutes of Health (R01CA251792).

\section{Data Availability}

The data that support the findings of this study are openly available at the following DOI: 10.7937/J47W-NM11. 

\clearpage

% following only if there is an appendix

\section*{References}
\addcontentsline{toc}{section}{\numberline{}References}
\vspace*{-20mm}

% Following assumes you are using bibtex. However, for submission to the
% journal you MUST explicitly INCLUDE THE REFERENCES IN THE TEX FILE. 
% In that case you need the following
%\begin{thebibliography}{10}
	
%\end{thebibliography}
%\begin{thebibliography}{10}
% insert the .bbl file generated by bibtex here
	%This will be a series of entries from your .bib file formatted
	%something like
	%\bibitem{Me09}
        %{I.~Meijsing, B.~W.~Raaymakers, A.~J.~E.~Raaijmakers \it et al.},
        %\newblock {Dosimetry for the MRI accelerator: the impact of a 
	%magnetic field on the response of a Farmer NE2571 ionization chamber},
        %\newblock Phys. Med. Biol. {\bf 54}, 2993 -- 3002 (2009).

%\end{thebibliography}

% The following is when using bibtex and picks up the example.bib file

%\bibliography{Explicit address of .bib file}
\bibliography{./bibliography}      %example.bib is on the same directory
% above points to where we find the master reference list
% and also causes the bibliography to be printed

% When creating your bibliography you should run bibtex on your local
% computer after running pdflatex on your .tex file. bibtex will
% generate a .bbl file.
% Copy the contents of this .bbl file into your main latex document,
% replacing the "\bibliography" command which was pointing at your .bib file.

% following defines style of .bbl file 

%\bibliographystyle{explicit relative path to medphy.bst}
\bibliographystyle{./medphy.bst}    %if this is installed on your system,
				    %it is not essential to have the    ./

% Note that you need to typeset once, then run bibtex, then typeset another
% two times to get the references working properly.
\section{Supplemental Materials}
\counterwithin{figure}{section}
\counterwithin{table}{section}
\setcounter{figure}{0}
\setcounter{table}{0}
\renewcommand{\thefigure}{S.\arabic{figure}}
\renewcommand{\thetable}{S.\arabic{table}}

\begin{table}[h!]
	\centering
	\begin{tabular}{c c c c}
		\hline
		\multirow{2}{*}{Image Parameter} & \multicolumn{3}{c}{Model} \\
		& Radiomics &	Foundation & ViT \\
		\hline
		Image Format &	Nifti &	Nifti &	Nifti \\
		Resampling &	None &	None & 	1x1x1 mm \\
		Cropping &	None &	Axial &	ROI-based \\
		Renormalization &	None &	None &	Min-Max \\
		Input Image &	\#x512x512 + GTVp &	\#x512x512 &	80x80x80 + GTVp \\
		Features to classifier &	1316 &	1408 &	256 \\
	\end{tabular}
	\caption{Summary of input differences between the radiomics, foundation and ViT features.}
	\label{tab:supp_input}
\end{table}
\clearpage

\begin{table}[h!]
	\centering
	\begin{tabular}{c}
		\hline
		Selected Radiomics Features \\
		\hline
		exponential\_gldm\_DependenceEntropy \\
		exponential\_gldm\_DependenceNonUniformityNormalized \\
		exponential\_glrlm\_GrayLevelNonUniformity \\
		exponential\_glrlm\_LongRunEmphasis \\
		exponential\_glrlm\_LongRunLowGrayLevelEmphasis \\
		gradient\_glrlm\_GrayLevelNonUniformity \\
		gradient\_ngtdm\_Busyness \\
		original\_glrlm\_GrayLevelNonUniformity \\
		original\_shape\_MajorAxisLength \\
		original\_shape\_Maximum2DDiameterColumn \\
		original\_shape\_Maximum2DDiameterRow \\
		original\_shape\_Maximum3DDiameter \\
		original\_shape\_SurfaceArea \\
		square\_gldm\_DependenceNonUniformityNormalized\\
		square\_gldm\_GrayLevelNonUniformity\\
		square\_glrlm\_GrayLevelNonUniformity\\
		square\_glrlm\_LongRunEmphasis\\
		square\_glrlm\_LongRunLowGrayLevelEmphasis\\
		square\_glrlm\_RunVariance\\
		squareroot\_glrlm\_GrayLevelNonUniformity\\
		wavelet-HHH\_gldm\_DependenceNonUniformityNormalized\\
		wavelet-HLH\_glrlm\_GrayLevelNonUniformity\\
		wavelet-HLL\_gldm\_GrayLevelNonUniformity\\
		wavelet-HLL\_glrlm\_GrayLevelNonUniformity\\
		wavelet-LHH\_glcm\_ClusterShade\\
		wavelet-LHL\_glcm\_Correlation\\
		wavelet-LLH\_gldm\_GrayLevelNonUniformity\\
		wavelet-LLH\_glrlm\_GrayLevelNonUniformity\\
		wavelet-LLL\_glcm\_SumSquares\\
		wavelet-LLL\_glrlm\_GrayLevelNonUniformity\\
		\hline
	\end{tabular}
	\caption{List of radiomics features selected by the mRMR algorithm for the SVM model.}
	\label{tab:supp_radiomics}
\end{table}

\begin{table}
	\centering
	\begin{tabular}{c}
		\hline
		Selected Foundation Embeddings \\
		\hline
		embedding\_11 \\
		embedding\_20\\
		embedding\_25\\
		embedding\_33\\
		embedding\_36\\
		embedding\_46\\
		embedding\_59\\
		embedding\_90\\
		embedding\_92\\
		embedding\_117\\
		embedding\_145\\
		embedding\_173\\
		embedding\_178\\
		embedding\_215\\
		embedding\_496\\
		embedding\_497\\
		embedding\_498\\
		embedding\_672\\
		embedding\_896\\
		embedding\_1096\\
		embedding\_1097\\
		embedding\_1098\\
		embedding\_1099\\
		embedding\_1100\\
		embedding\_1101\\
		embedding\_1102\\
		embedding\_1103\\
		embedding\_1184\\
		embedding\_1185\\
		embedding\_1272\\
		\hline
	\end{tabular}
	\caption{List of foundation embeddings selected by the mRMR algorithm for the SVM model. The numbers correspond to the index position of the embeddings as they are output from the CT Foundation API.}
	\label{tab:supp_foundation}
\end{table}

\begin{table}
	\centering
	\begin{tabular}{c c}
		\hline
		Parameter &	Grid Search Range \\
		\hline
		mRMR N features &	[3,5,10,15,20,30,50,100,500] \\
		Smote k neighbors &	[3, 5, 10, 20] \\
		SVM C &	[0.1, 1, 10] \\
		SVM kernel &	[linear, rbf, poly] \\
		SVM N degrees &	[2,3,4] \\
		SVM gamma &	[0.01, 0.1, 1, 10, scale] \\
		\hline
	\end{tabular}
	\caption{The grid search parameters for the traditional machine learning models. Separate grid searches were done for the SVM parameters.}
	\label{tab:supp_grid}
\end{table}

\begin{figure}
	\centering
	\begin{subfigure}[b]{0.45\textwidth}
		\centering
		\includegraphics[width=\textwidth]{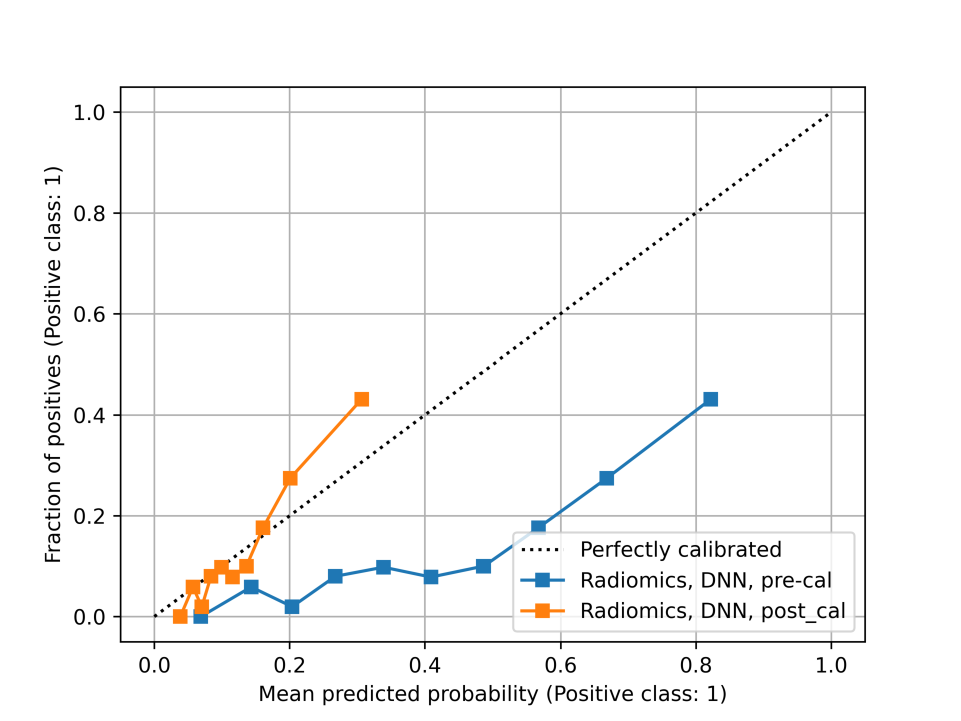}
		\caption{}\label{subfig:s1a}
	\end{subfigure}
	\begin{subfigure}[b]{0.45\textwidth}
		\centering
		\includegraphics[width=\textwidth]{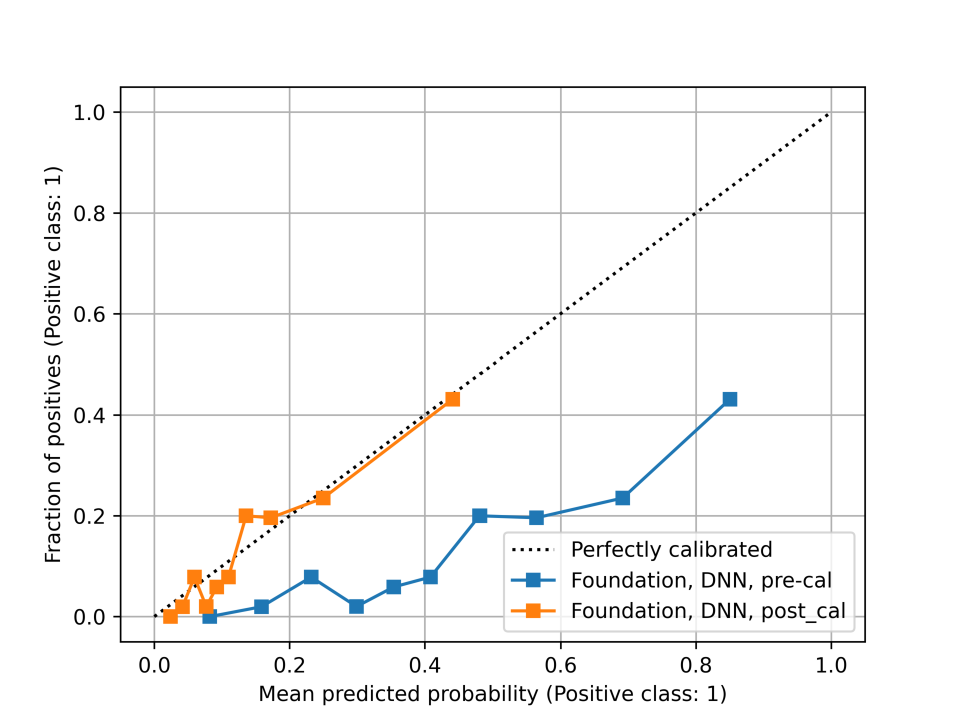}
		\caption{}\label{subfig:s1b}
	\end{subfigure}
	\begin{subfigure}[b]{0.45\textwidth}
		\centering
		\includegraphics[width=\textwidth]{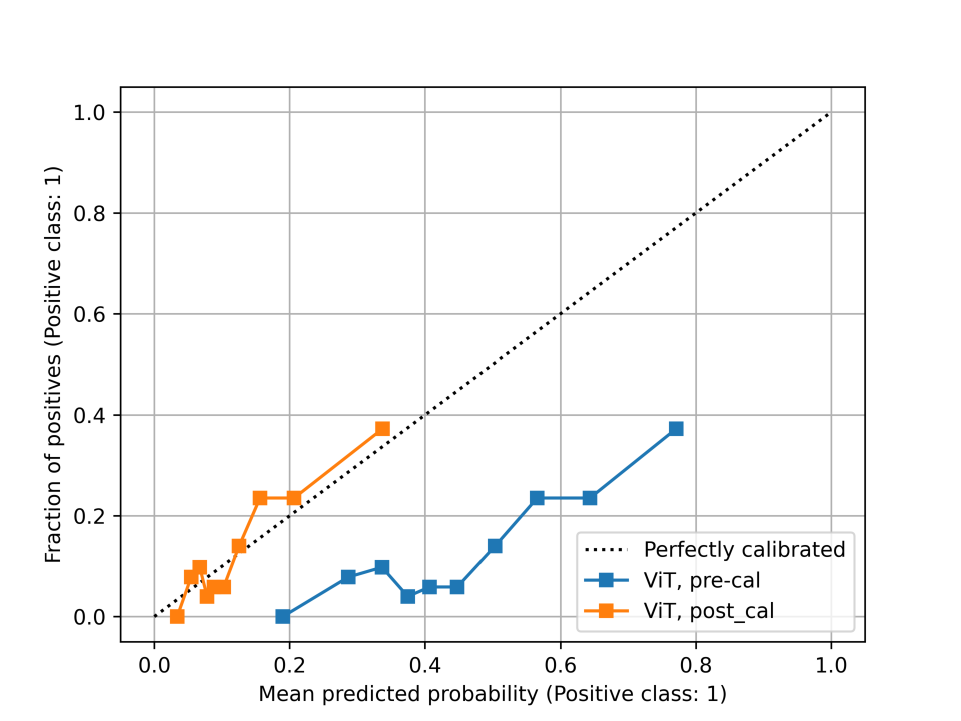}
		\caption{}\label{subfig:s1c}
	\end{subfigure}
	\begin{subfigure}[b]{0.45\textwidth}
		\centering
		\includegraphics[width=\textwidth]{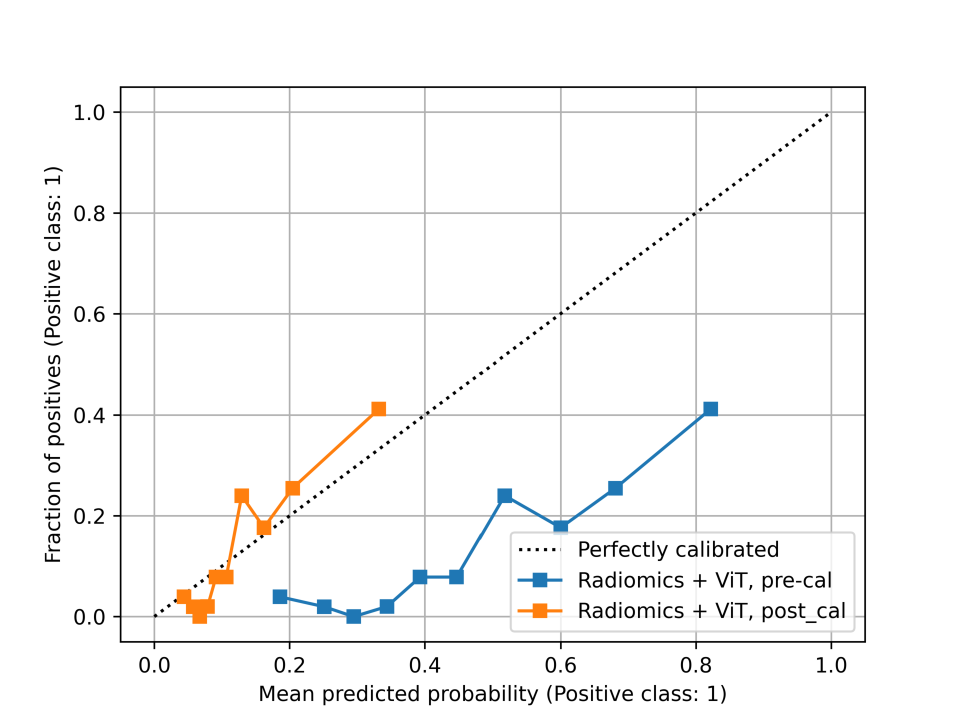}
		\caption{}\label{subfig:s1d}
	\end{subfigure}
	\caption{Calibration curves for the DNN models before and after calibration.}
	\label{fig:supp_calibration_dnn}
\end{figure}

\begin{figure}
	\centering
	\begin{subfigure}[b]{0.45\textwidth}
		\centering
		\includegraphics[width=\textwidth]{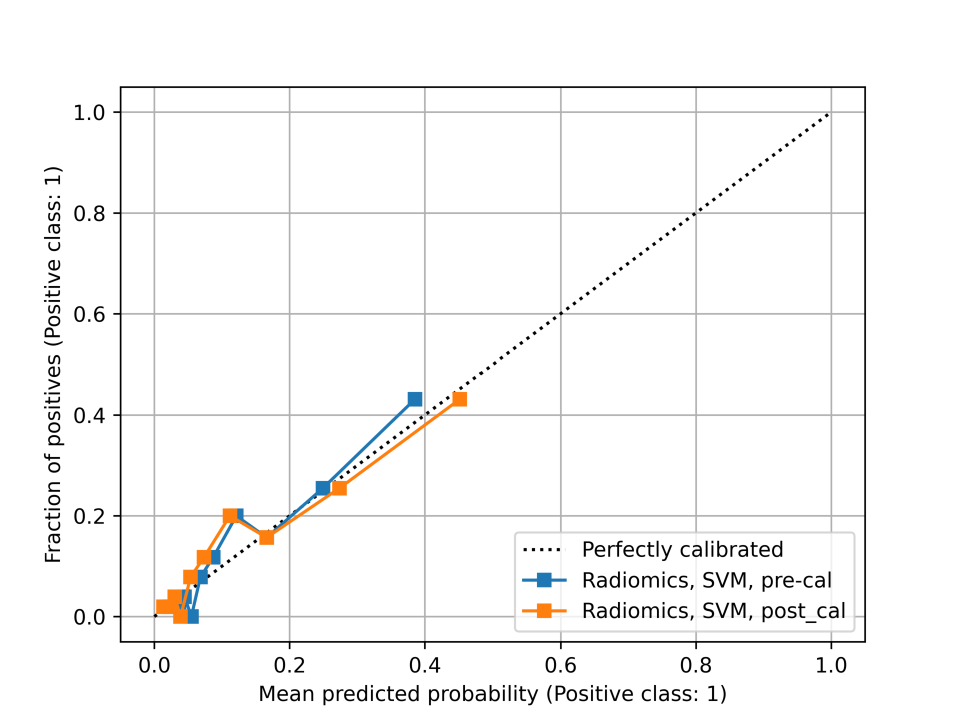}
		\caption{}\label{subfig:s2a}
	\end{subfigure}
	\begin{subfigure}[b]{0.45\textwidth}
		\centering
		\includegraphics[width=\textwidth]{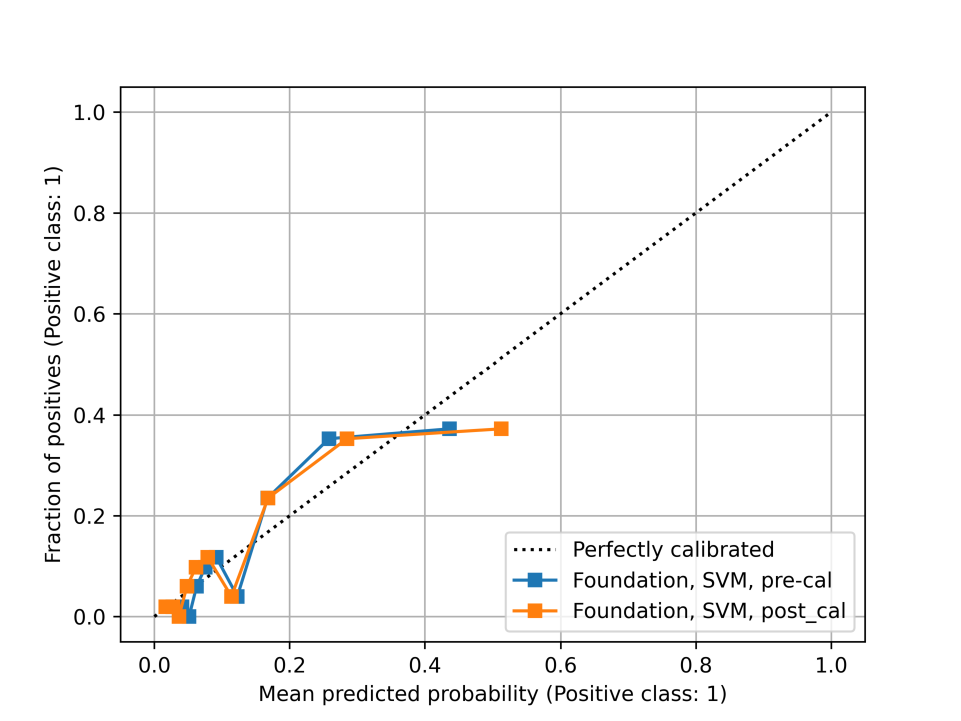}
		\caption{}\label{subfig:s2b}
	\end{subfigure}
	\begin{subfigure}[b]{0.45\textwidth}
		\centering
		\includegraphics[width=\textwidth]{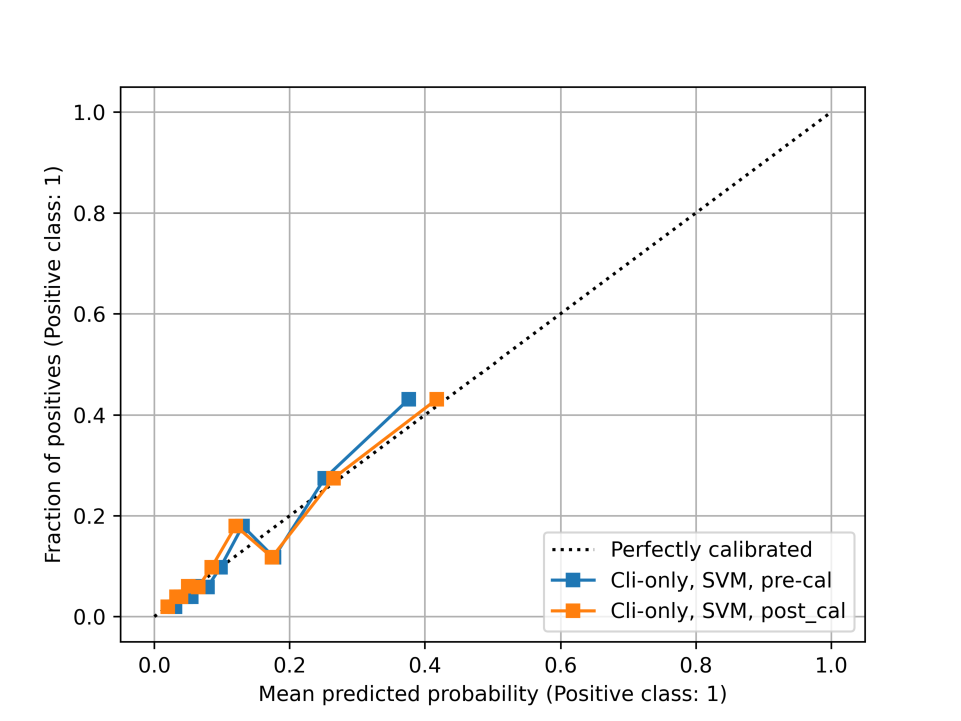}
		\caption{}\label{subfig:s2c}
	\end{subfigure}
	\caption{Calibration curves for the SVM models before and after calibration.}
	\label{fig:supp_calibration_trad}
\end{figure}

\begin{table}
	\centering
	\begin{tabular}{c c}
		\hline
		Model &	Threshold \\
		\hline
		Radiomics, DNN &	0.139 \\
		Foundation, DNN &	0.126 \\
		Image, DNN &	0.128 \\
		Image+Radiomics, DNN &	0.123 \\
		Radiomics, SVM &	0.060 \\
		Foundation, SVM &	0.064 \\
		No cli, SVM &	0.073 \\
		\hline
	\end{tabular}
	\caption{Retuned thresholds for the various models after performing a calibration.}
	\label{tab:supp_threshold}
\end{table}

\begin{table}
	\centering
	\begin{tabular}{c c c c c c}
		\hline
		Model &	CV Fold &	AUC [95\% CI] &	AP &	SEN &	SPE \\
		\hline
		\multirow{6}{*}{Foundation} &	0 &	0.784 [0.727, 0.841] &	0.426 &	0.657 &	0.755 \\
		&1&	0.776 [0.718, 0.834]&	0.417&	0.657&	0.757 \\
		&2&	0.789 [0.733, 0.844]&	0.412&	0.687&	0.717 \\
		&3&	0.789 [0.732, 0.845]&	0.421&	0.687&	0.753 \\
		&4&	0.78 [0.724, 0.837]&	0.379&	0.806&	0.669 \\
		&Avg ±std&	0.784 ± 0.006&	0.411 ± 0.019&	0.699 ± 0.062&	0.730 ±0.038 \\
		\multirow{6}{*}{Radiomics}&	0&	0.78 [0.723, 0.837]&	0.343&	0.836&	0.558 \\
		&1&	0.754 [0.692, 0.816]&	0.357&	0.552&	0.791 \\
		&2&	0.779 [0.721, 0.838]&	0.379&	0.821&	0.540 \\
		&3&	0.763 [0.703, 0.823]&	0.387&	0.627&	0.760 \\
		&4&	0.739 [0.672, 0.806]&	0.359&	0.582&	0.741 \\
		&Avg ±std&	0.763 ± 0.017&	0.365 ± 0.019&	0.684 ± 0.135&	0.678 ± 0.119 \\
		\multirow{6}{*}{ViT}&	0&	0.741 [0.676, 0.806]&	0.344&	0.657&	0.723 \\
		&1&	0.727 [0.661, 0.792]&	0.332&	0.672&	0.683 \\
		&2&	0.756 [0.693, 0.819]&	0.382&	0.701&	0.735 \\
		&3&	0.757 [0.693, 0.82]&	0.358&	0.731&	0.644 \\
		&4&	0.757 [0.693, 0.822]&	0.366&	0.716&	0.707 \\
		&Avg ±std&	0.748 ± 0.013&	0.356 ± 0.019&	0.695 ± 0.030&	0.698 ± 0.036 \\
		\multirow{6}{*}{Radiomics + ViT}&	0&	0.781 [0.722, 0.841]&	0.367&	0.701&	0.739 \\
		&1&	0.8 [0.747, 0.853]&	0.398&	0.896&	0.562 \\
		&2&	0.799 [0.747, 0.85]&	0.370&	0.896&	0.583 \\
		&3&	0.758 [0.694, 0.821]&	0.374&	0.657&	0.728 \\
		&4&	0.777 [0.719, 0.835]&	0.375&	0.731&	0.687 \\
		&Avg ±std&	0.783 ± 0.017&	0.377 ± 0.012&	0.776 ± 0.112&	0.660 ± 0.082 \\
		\hline
	\end{tabular}
	\caption{Performance of the testing set on the selected DNN models from each cross-validation fold, before applying calibration. The AUC, average precision, sensitivity, and specificity are provided}
	\label{tab:supp_folds}
\end{table}

\begin{table}
	\centering
	\begin{tabular}{c c c c c}
		Model + Selection&	AUC [95\%CI]&	AP &	SEN&	SPE \\
		\hline
		\multicolumn{5}{c}{Foundation Models} \\
		\hline
		Foundation + Contrast&	0.780&	0.372&	0.750&	0.685 \\
		Foundation + No Contrast&	0.847&	0.667&	0.909&	0.682 \\
		Foundation + Male&	0.790&	0.437&	0.780&	0.668 \\
		Foundation + Female&	0.786&	0.238&	0.750&	0.763 \\
		Foundation + No clinical&	0.707&	0.350&	0.687&	0.562 \\
		Foundation&	0.791& 	0.412&	0.776&	0.685 \\
		\hline
		\multicolumn{5}{c}{Radiomics Models} \\
		\hline
		Radiomics + Contrast&	0.762&	0.334&	0.679&	0.712 \\
		Radiomics + No Contrast&	0.822&	0.617&	0.727&	0.750 \\
		Radiomics + Male&	0.770&	0.392&	0.678&	0.704 \\
		Radiomics + Female&	0.766&	0.318&	0.750&	0.750 \\
		Radiomics + No clinical&	0.718&	0.261&	0.642&	0.644 \\
		Radiomics&	0.772 &	0.379&	0.687&	0.712 \\
		\hline
	\end{tabular}
	\caption{Performance of two DNN models using radiomics and foundation embeddings as input for different selections of patient populations and for separately trained models that remove the use of clinical features. The original model performances are provided for comparison.}
	\label{tab:supp_strat}
\end{table}

\end{document}